\documentclass{ieeeaccess}
\usepackage{cite}
\usepackage{algorithm}  
\usepackage{algorithmic}
\usepackage{amsmath,amssymb,amsfonts}

\usepackage[caption=false,font=normalsize,labelfont=sf,textfont=sf]{subfig}
\usepackage{textcomp}
\usepackage{stfloats}
\usepackage{url}
\usepackage{verbatim}
\usepackage{graphicx}
\usepackage{threeparttable}
\usepackage{subfig}
\usepackage{cite}
\usepackage{textcomp}
\usepackage{threeparttable}
\def\BibTeX{{\rm B\kern-.05em{\sc i\kern-.025em b}\kern-.08em
    T\kern-.1667em\lower.7ex\hbox{E}\kern-.125emX}}
\begin{document}
\history{Date of publication xxxx 00, 0000, date of current version xxxx 00, 0000.}
\doi{10.1109/ACCESS.2017.DOI}

\title{CNTS: Cooperative Network for Time Series}

\author{\uppercase{Jinsheng Yang}\authorrefmark{1},
\uppercase{Yuanhai Shao\authorrefmark{1}},
\uppercase{ChunNa Li\authorrefmark{1}}.}
\address[1]{Management School, Hainan University, Haikou, 570228, P.R.China}

\tfootnote{This work is supported in part by National Natural Science Foundation of China (Nos. 12271131, 61966024, 62106112, 61866010 and 11871183), in part by the Natural Science Foundation of Hainan Province (No.120RC449).}

\markboth
{Author \headeretal: Preparation of Papers for IEEE TRANSACTIONS and JOURNALS}
{Author \headeretal: Preparation of Papers for IEEE TRANSACTIONS and JOURNALS}

\corresp{Corresponding author: Yuanhai Shao (e-mail: shaoyuanhai21@163.com).}

\begin{abstract}
The use of deep learning techniques in detecting anomalies in time series data has been an active area of research with a long history of development and a variety of approaches. In particular, reconstruction-based unsupervised anomaly detection methods have gained popularity due to their intuitive assumptions and low computational requirements. However, these methods are often susceptible to outliers and do not effectively model anomalies, leading to suboptimal results. This paper presents a novel approach for unsupervised anomaly detection, called the Cooperative Network Time Series (CNTS) approach. The CNTS system consists of two components: a detector and a reconstructor. The detector is responsible for directly detecting anomalies, while the reconstructor provides reconstruction information to the detector and updates its learning based on anomalous information received from the detector. The central aspect of CNTS is a multi-objective optimization problem, which is solved through a cooperative solution strategy. Experiments on three real-world datasets demonstrate the state-of-the-art performance of CNTS and confirm the cooperative effectiveness of the detector and reconstructor. The source code for this study is publicly available on GitHub \footnote{https://github.com/BomBooooo/CNTS/tree/main}.
\end{abstract}


\titlepgskip=-15pt

\maketitle

\section{Introduction}
\label{sec:introduction}
\IEEEPARstart{T}{he} field of anomaly detection has a rich history dating back to the 1960s \cite{grubbs1969procedures}, with an increasing number of applications in areas such as fraud detection \cite{phua2010comprehensive} and intrusion detection \cite{shone2018deep}. The proliferation of data has further expanded the scope of anomaly detection. The presence of anomalies in Internet of Things (IoT) data poses significant security risks, hindering the development of the IoT \cite{liu2020deep}. Deep Learning, a potent machine learning technique, has recently garnered significant attention from researchers and demonstrated its superiority in anomaly detection \cite{pang2021deep}.

The problem of anomalies is very common in real life, but it is difficult to obtain high quality labeled data \cite{cook2019anomaly}. Therefore, there is an urgent need for unsupervised anomaly detection methods. Deep Learning approaches for anomaly detection can broadly be classified into three categories \cite{pang2021deep}: ((\romannumeral1) Feature extraction: the objective is to use Deep Learning to reduce high-dimensional and/or non-linearly separable data into low-dimensional feature representations for downstream anomaly detection \cite{xu2015learning}. (\romannumeral2) Learning feature representations of normality: these methods integrate feature learning with anomaly assessment, unlike the previous category which separates the two processes. Examples include autoencoders \cite{kieu2019outlier}, GANs \cite{li2021dct}, predictability modeling \cite{chen2021learning}, self-supervised classification \cite{wang2019effective}. (\romannumeral3) End-to-end Anomaly Score Learning: the goal is to learn scalar anomaly scores in a direct manner without relying on existing anomaly metrics. The neural network in this approach learns to identify outliers directly \cite{sabokrou2018adversarially}. The first two methods of anomaly detection are predominantly utilized in unsupervised tasks, whereas the third method is more commonly employed in supervised tasks, as it necessitates the availability of anomalous labels. 

The first category of methods involves a separation of tasks and models, while the third category requires labeled abnormal data. As a result, the most commonly used method for unsupervised anomaly detection currently belongs to the second category, i.e. reconstruction-based methods \cite{ruff2021unifying}. These method operates by inputting a time series data and calculating its reconstruction error, which is then used as an anomaly score to determine the abnormality of the data. Once the anomaly score has been obtained, a suitable threshold is selected to identify anomalies \cite{yin2020anomaly}. The underlying assumption of this method is that deep networks can effectively learn low-dimensional features of the data and normalize outliers by reconstruction. 

Although these methods have demonstrated good performance, they suffer from two key limitations. Firstly, they do not provide any special treatment for outliers, which can negatively impact the results of the model as they are present throughout the training process. Secondly, the goal of the network is reconstruction, rather than anomaly detection, resulting in suboptimal performance \cite{pang2021deep}.

In this paper, we present a cooperative network for time series (CNTS) to address the aforementioned limitations of existing reconstruction-based methods. The CNTS consists of two deep learning networks, a reconstructor (\textbf{R}) and a detector (\textbf{D}). \textbf{R} is responsible for reconstructing the original data and providing abnormal labels by calculating reconstruction error, while \textbf{D} trains based on these labels and returns the results to \textbf{R} for weight adjustment. Through this cooperative training process, \textbf{D} and \textbf{R} work together to improve learning performance. The CNTS reduces the impact of outliers by screening and selecting appropriate data during training, and enhances anomaly detection performance by separating the tasks of data reconstruction and anomaly detection into separate networks. The CNTS solves a multiobjective programming problem and is based on the FEDFormer network \cite{zhou2022fedformer}. The main contributions of this work can be summarized as follows:

\begin{itemize}
	\item {\texttt{Reconstruction}}: a novel reconstruction network, based on anomaly weights, is proposed in this work. This reconstruction network has the ability to adapt its own training loss to the abnormal conditions, thus enhancing its ability to distinguish between abnormal and normal samples after reconstruction.
	\item {\texttt{Detection}}: we present a direct training method for unsupervised anomaly detection by separating the tasks of reconstruction and anomaly detection. This approach avoids suboptimal results caused by suboptimal feature extraction.
	\item {\texttt{Cooperation Network}}: A new cooperative training mode for the network is proposed in this work. The network consists of two parts: the reconstructor (\textbf{R}) that learns the characteristics of normal samples and the detector (\textbf{D}) that detects anomalies. \textbf{R} and \textbf{D} work together and enhance each other's performance. Both are part of each other's loss function in this cooperative training mode.
	\item {\texttt{Experiment}}: Experiments were conducted on three databases, totaling 128 real-world data, and the results showed that the detection performance of \textbf{D} surpasses that of nine baseline methods in terms of accuracy, F1 value, and AUC. Additionally, the ability to distinguish between normal and abnormal samples was further improved by the reconstructed data of \textbf{R}.
\end{itemize}

\section{Problem Formulate}
Given a period of univariate time series data $ \bold{X} = \{\bold{x_1}, \bold{x_2}, \dots, \bold{x_N} \} $, with $ \textbf{x}_t \in \mathbb{R}^1 $, where $ N $ is the length of data. The abnormal labels are denoted as $ \bold{Y} = \{ y_1, y_2, \dots, y_N \}$, where 
\begin{equation}
	y_t =
	\begin{cases}
		1 \quad if\ \bold{x_t}\ is\ abnormal,\\
		0 \quad else.
	\end{cases}
\end{equation}

Therefore, the anomaly detection problem can be described in the following form: Given a period of time series data  $ \bold{X} = \{\bold{x_1}, \bold{x_2}, \dots, \bold{x_N} \} $, the following objective function is achieved by training the model
\begin{equation}
	\underset{\theta_{\mathcal{M}}}{min}\ \mathcal{L}(\mathcal{M}(\bold{X}, \theta_{\mathcal{M}}), \bold{Y}).
	\label{anormaly_detection}
\end{equation}
where $\mathcal{L}$ represents the loss function and $\mathcal{M}(\cdot, \theta_{\mathcal{M}})$ represents the model $\mathcal{M}$ with parameter $\theta_{\mathcal{M}}$. Since this paper mainly discusses the problem of unsupervised anomaly detection, $\bold{Y}$ does not exist in the training set and only exists in the testing set, which is same as the usual unsupervised anomaly detection settings. 

\section{Ralated works}
The objective of reconstruction-based anomaly detection models is to learn an optimized model that accurately reconstructs normal data instances while failing to reconstruct anomalies effectively \cite{ruff2021unifying}. The use of reconstruction models for anomaly detection is based on the assumption that the reconstructed data can differentiate normal data from abnormal data. The objective function of these models can be formulated as follows:

\begin{equation}
	\underset{\theta_{\mathcal{M}}}{min}\ \sum_{i=1}^{n}[(1 - y_i)\mathcal{L}(\bold{x_i}, \mathcal{M}(\bold{x_i}, \theta_{\mathcal{M}})) - y_i \mathcal{L}(\bold{x_i}, \mathcal{M}(\bold{x_i}, \theta_{\mathcal{M}}))]
	\label{objective}
\end{equation}

The effectiveness of the reconstruction model in detecting anomalies depends on its ability to produce a small reconstruction error for normal values and a large reconstruction error for abnormal values. This will result in a relatively low value for formula (\ref{objective}). On the other hand, formula (\ref{anormaly_detection}) has an advantage over formula (\ref{objective}) as it models anomalies directly, whereas formula (\ref{objective}) relies on the hope that the reconstruction error can differentiate between normal and abnormal values.

In practical modeling, the labels for anomalies are always unknown, and the objective function is formulated as equation (\ref{rec}). Equation (\ref{rec}) simplifies the second term in equation (\ref{objective}). Despite the fact that outliers only represent a small fraction of the original data, equation (\ref{rec}) remains effective.

\begin{equation}
	\underset{\theta_{\mathcal{M}}}{min}\ \frac{1}{n} \sum_{i=1}^{n}\vert \vert \bold{x_i} - \mathcal{M}(\bold{x_i}, \theta_{\mathcal{M}}) \vert \vert ^2
	\label{rec}
\end{equation}

There are already several models that are based on these concepts. For instance, the Deep Autoencoding Gaussian Mixture Model (DAGMM) \cite{zong2018deep} is an unsupervised anomaly detection method that combines an autoencoder with a Gaussian mixture model. The autoencoder is used to compress and reconstruct the data, and the reconstruction error is fed into the Gaussian mixture model. The entire model is trained end-to-end. MAD-GAN \cite{li2019mad} is a model that employs a Generative Adversarial Network (GAN) architecture for training, and uses both the reconstruction loss and the discriminative loss as anomaly scores in the testing phase. The Unsupervised Anomaly Detection (USAD) \cite{audibert2020usad} method introduces another decoder based on the traditional autoencoder, and trains the model unsupervised through adversarial learning. The reconstruction errors from the two decoders play a crucial role in the training process. MTAD-GAT \cite{zhao2020multivariate} optimizes both the forecast-based model and the reconstruction-based model and achieves a better representation of the time series by combining single-timestamp forecasting and the reconstruction of the entire time series. CAE-M \cite{zhang2021unsupervised} simultaneously optimizes two networks, the characterization network, and the memory network, where the characterization network is based on the reconstruction loss.

Methods such as these can partially alleviate the impact of outliers because normal values typically constitute the majority of the data. However, it is not feasible to use the same weights for all data, as outliers can affect the training process. Additionally, the lack of direct training for anomalies in the objective function can result in suboptimal outcomes.

The inspiration for the cooperative approach in this paper draws mainly from the adversarial process in GAN \cite{goodfellow2020generative}. As is well-known, the original GAN was primarily used for training generative models, and various modifications have since been developed. In GANs, the generator and discriminator engage in a game and eventually reach a Nash equilibrium. The patterns generated by the generator in the equilibrium state are sufficiently similar to real patterns to confuse the discriminator. This paper introduces a Cooperative Network for Time Series (CNTS), which also contains two networks: a reconstructor and a detector. However, the relationship between the two networks is not adversarial, but cooperative. The reconstructor (\textbf{R}) and detector (\textbf{D}) work together towards similar objectives, ultimately achieving their individual goals. Specifically, \textbf{R}'s goal is to reconstruct normal samples and distinguish them from abnormal samples, while \textbf{D}'s goal is to detect abnormalities. There is a degree of overlap between the two goals, allowing for mutual reinforcement.

\section{Cooperative Network for Time Series}
An anomaly is a singular or series of observations that significantly deviate from the general distribution of the data, with the set of anomalies constituting a small portion of the data set \cite{braei2020anomaly}. Therefore, anomaly detection can be viewed as identifying values that do not conform to the known data distribution. To address this challenge, we propose a deep learning network that integrates reconstruction and detection. Reconstruction aims to reconstruct values that closely match the known data distribution, while detection focuses on identifying values that do not align with the fitted data distribution. These two tasks are both distinct and interrelated.

\textbf{Difference}: There are two main differences between the objectives of the reconstructor (\textbf{R}) and the detector (\textbf{D}). Firstly, \textbf{R} aims to fit the known data distribution, while \textbf{D} aims to identify values that deviate from the normal data distribution. Secondly, the output information differs, with \textbf{R} providing reconstruction and \textbf{D} providing anomaly scores.

\textbf{Similarity}: Despite the differences in their objectives, \textbf{R} and \textbf{D} share similarities in their input forms, and the output information can be used to some extent to judge the degree of abnormality.

So, the mutual utilization of information between two entities, \textbf{D} and \textbf{R}, has the following advantages. The reconstruction error of \textbf{R} can serve as a labeling mechanism for \textbf{D} during its training process. On the other hand, \textbf{D} can provide information to \textbf{R} regarding anomalies, reducing the influence of potential outliers on \textbf{R}'s training. This separation of the task of anomaly detection from data reconstruction enables CNTS to be modeled as a multi-objective programming problem. The objective function is formulated as follows:
\begin{equation}
	\left [ 
	\begin{aligned}
		&\underset{\theta_{\bold{D}}, \theta_{\bold{R}}}{min}\ \sum_{i=1}^{n} \mathcal{L}_D(\bold{D}(\bold{x_i}, \theta_{\bold{D}}), \mathcal{L}_R(\bold{x_i}, \bold{R}(\bold{x_i}, \theta_{\bold{R}})))\\
		&\underset{\theta_{\bold{D}}, \theta_{\bold{R}}}{min}\ \sum_{i=1}^{n}(1 - \hat{y_i}(\bold{x_i}, \theta_{\bold{D}}))\mathcal{L}_R(\bold{x_i}, \bold{R}(\bold{x_i}, \theta_{\bold{R}}))
	\end{aligned}
	\right ]  \\
	\label{CNTS}
\end{equation}
\begin{equation*}
	s.t.\ \hat{y_t}(\bold{x_i}, \theta_{\bold{D}}) =
	\begin{cases}
		1 \quad if\ \bold{D}(\bold{x_i}, \theta_{\bold{D}}) > Top_{k\%}(\bold{D}(\bold{X}, \theta_{\bold{D}})),\\
		0 \quad else.
	\end{cases}
\end{equation*}
where $ \bold{D}(\cdot, \theta_D) $ is the \textbf{D} with parameters $ \theta_D $, $ \bold{R}(\cdot, \theta_R) $ is the \textbf{R} with parameters $ \theta_R $, $\mathcal{L}_D$ and $\mathcal{L}_R$ denote the loss of \textbf{D} and \textbf{R} respectively, and $\hat{y_t}(\cdot, \theta_{\bold{D}})$ is a category label, if the value of $ \bold{D}(\bold{x_i}, \theta_{\bold{D}}) $ accounts for the top $ k\% $ of all $ \bold{D}(\bold{X}, \theta_{\bold{D}}) $, then $\hat{y_t}(\bold{x_i}, \theta_{\bold{D}})$ equals to 1, otherwise it equals to 0. Through this label, the data with a large abnormal score is screened out during the training process, thereby reducing the impact of outliers on the reconstruction model.

To optimize the performance of the proposed CNTS, it is essential to enhance the reconstruction effect of \textbf{R} and make the features of outliers in the reconstruction error more evident. During the training process, while the numerical value of the abnormal reconstruction error may not necessarily increase, it should increase relative to the reconstruction error of the normal points. At the same time, \textbf{D} is continually improving. The improvement of \textbf{D} is achieved by using the reconstruction error provided by \textbf{R} as a criterion for detecting anomalies. As \textbf{D} becomes more robust, the anomaly detection is passed back to \textbf{R}, which results in a clearer distinction between normal and anomalous reconstructions. The following sections describe a highly effective cooperative learning strategy for optimizing both objectives simultaneously.

\subsection{Basic Model}
In this study, both the anomaly detector and the reconstructor are based on the FEDFormer \cite{zhou2022fedformer} network structure. The FEDFormer architecture integrates the Transformer model with the seasonal-trend decomposition method, leading to improved performance for long-term time series forecasting. This approach leverages the fact that many time series data can be represented sparsely in well-known bases and augments the Transformer with a frequency-based approach. The FEDFormer method has been shown to produce a reduction in prediction error of 14.8\% for multivariate and 22.6\% for univariate time series, compared to state-of-the-art methods, while being more computationally efficient than standard Transformers. As such, the FEDFormer structure serves as the foundation of the CNTS model in this paper.

\subsection{Anormaly Detection}

\begin{figure*}[!htb]
	\centering
	\includegraphics[width=\linewidth,trim=5 130 5 130,clip]{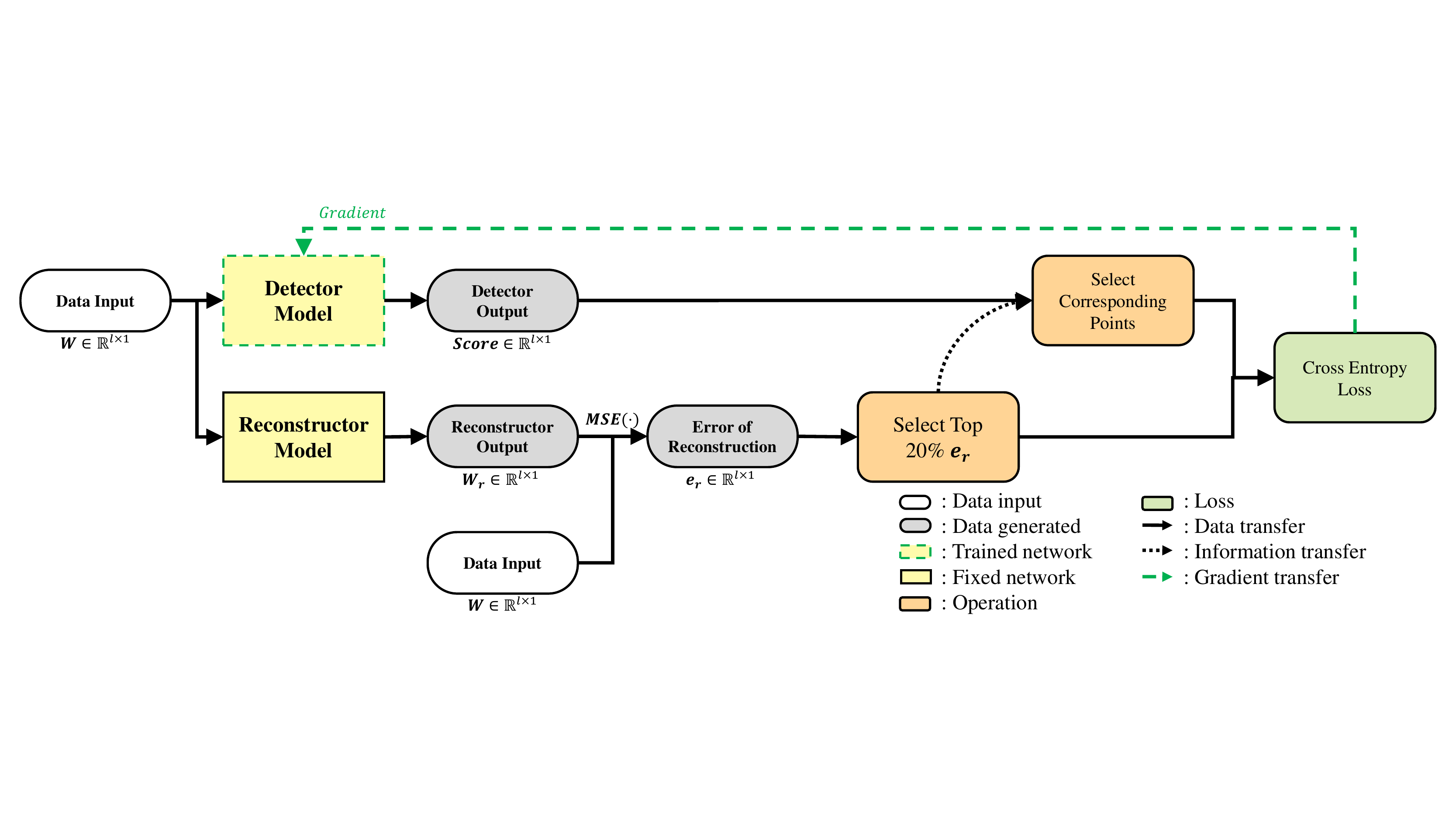}
	\caption{Train phase of \textbf{D}. During the training of \textbf{D}, the parameters of \textbf{R} are fixed. }
	\label{train_d}
\end{figure*}

The specific training process is shown in Figure \ref{train_d}. In the training of deep learning, before the data is input into the model, it needs to be cut through a sliding window of length $ \bold l $, denoted as $ \bold{W} $, where $ \bold{W} = \{ \bold{w_1}, \bold{w_2}, \dots, \bold{w_n} \} $, and $ \bold{w_i} = \{ \bold{x_i}, \bold{x_{i+1}}, \dots, \bold{x_{i+l-1}} \} $. 

In the training of \textbf{D}, it is assumed that data with larger reconstruction errors are more likely to be anomalies. The data entered into the model follows two paths. The first path involves inputting the data into \textbf{R} to obtain the reconstructed window $\bold{W_r}$ and calculating the reconstruction error $\bold{e_r}$. The second path involves inputting the data into \textbf{D} to obtain the abnormal score $Score$. Prior to the convergence of the two paths to calculate the loss, the outputs of both parts are screened, and the data that is most likely to be abnormal is selected for training the model. During the training of \textbf{D}, the parameters of \textbf{R} remain fixed.

The $\bold{W_r}$ and $\bold{e_r}$ can be calculated as
\begin{equation}
	\bold{W_r}(\bold{W}, \theta_R) = \bold{R}(\bold{W, \theta_R)}
\end{equation}
\begin{equation}
	\bold{e_r}(\bold{W}, \theta_R) = \mathcal{L}_M(\ \bold{W_r}(\bold{W}, \theta_R),\ \bold{W})
\end{equation}
where $ \mathcal{L}_M(\cdot, \cdot) $ is the mean squared error (MSE) loss function, $ \bold{R}(\cdot, \theta_R) $ is the \textbf{R} with parameters $ \theta_R $.  The abnormal score $\bold{Score}$ is calculated by 
\begin{equation}
	{Score}(\bold{W}, \theta_D) =\bold{D}(\bold{W}, \theta_D).
\end{equation}

To avoid the training collapse that can result from class imbalance, only the $Score$ values corresponding to the largest 20\% of points are selected as training points. The objective function can be expressed as follows:
\begin{equation}
	\begin{aligned}
		\theta_D^* = \arg \underset{\theta_D} {min}\ \mathcal{L}_C(&Softmax(\bold{e_{r_s}}(\bold{W}, \theta_R)), \\ &Softmax({{Score}_s}(\bold{W}, \theta_D)))
	\end{aligned}
\end{equation}
where $\mathcal{L}_C$ is CrossEntropy loss, and $ \bold{e_{r_s}} $ and $ {Score}_s $ represent selected $ \bold{e_r} $ and $ {Score} $, respectively. 

\subsection{Reconstruction}
\begin{figure*}[!htb]
	\centering
	\includegraphics[width=\linewidth,trim=20 120 20 120,clip]{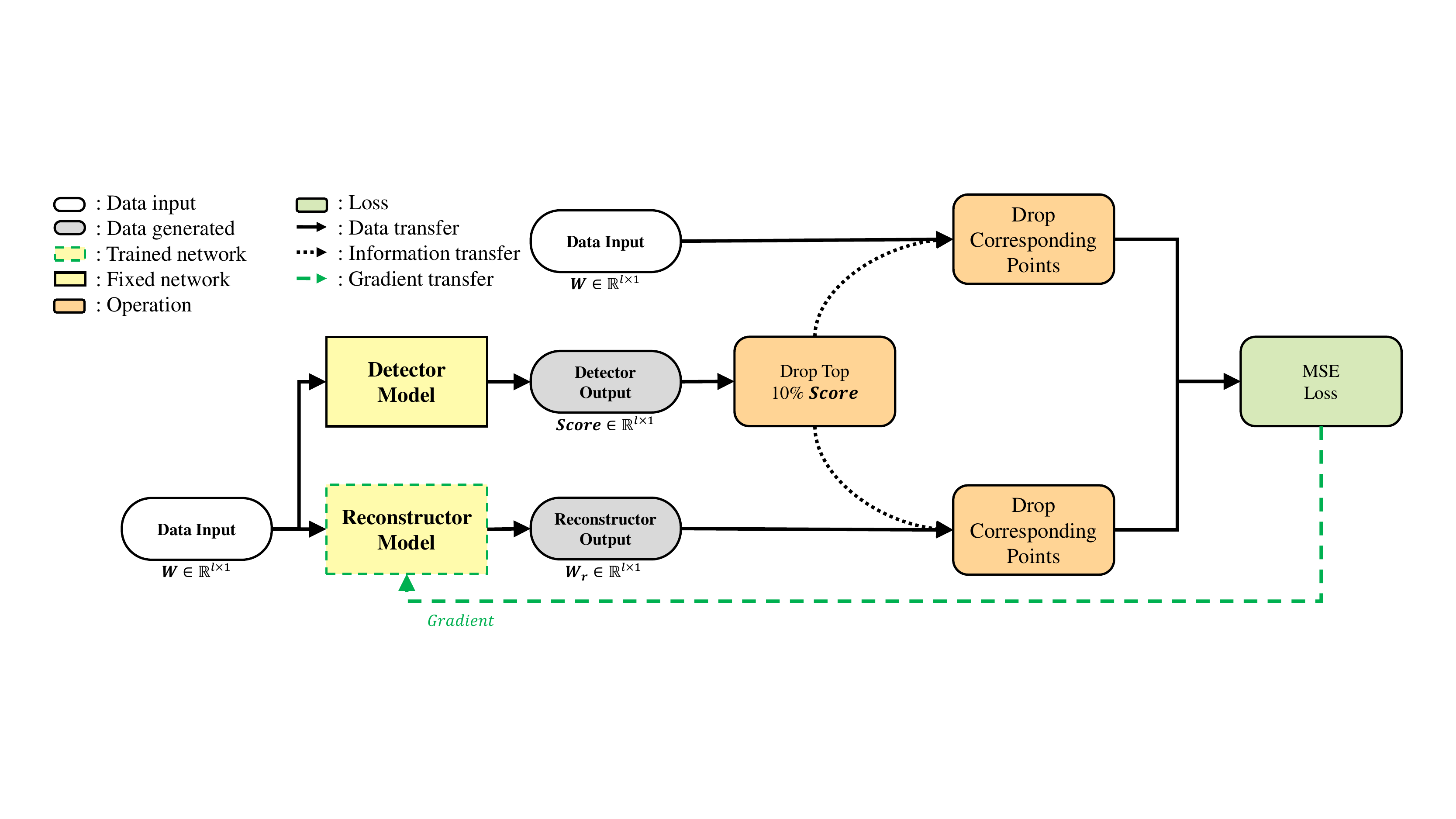}
	\caption{Train phase of \textbf{R}. During the training of \textbf{R}, the parameters of \textbf{D} are fixed.}
	\label{train_r}
\end{figure*}

Figure \ref{train_r} depicts the training process of \textbf{R}. It is assumed that the detector, trained by the method outlined in the previous section, has the capability to detect outliers. The input data must first pass through \textbf{D} to obtain the abnormal scores, and then it is processed by \textbf{R} to generate the reconstructed result. Prior to computing the loss between the input data and the reconstructed data, it is necessary to exclude the values with high abnormal scores from the data, as identified by the detector. The loss is then calculated by combining the input data and reconstructed data, and this guides the training of \textbf{R}. During the training of \textbf{R}, the parameters of \textbf{D} are held constant.

To mitigate the impact of outliers on the reconstruction process, \textbf{R} employs a strategy of excluding the top 10\% of values with the highest abnormal scores, as determined by \textbf{D}. This reduces the effect of outliers on the calculation of the loss, which is used to guide the training of \textbf{R}. The objective function of \textbf{R} is to minimize this calculated loss between the input and reconstructed data.
\begin{equation}
	\theta_R^* = \arg \underset{\theta_R} {min}\ \mathcal{L}_M(\bold{W}_s ,\ \bold{W}_{r_s}(\bold{W}, \theta_R))
\end{equation}
where $\theta_R$ is the parameter of \textbf{R}, $\bold{W}_s$ and $\bold{W}_{r_s}$ represent the selected $\bold{W}$ and $\bold{W}_{r}$. 

\subsection{Algorithm}

This subsection provides the pseudocode for the complete CNTS algorithm, referred to as Algorithm \ref{alg1}. In the algorithm, where $MSE(\bold{X_1}, \bold{X_2}) = (\bold{X_1} - \bold{X_2}) ^ 2$, the term "epoch" represents the number of iterations of the entire algorithm. "R\_epochs" and "D\_epochs" denote the number of iterations of \textbf{R} and \textbf{D}, respectively, in each loop of the algorithm. During the training of one model, the parameters of the other model remain fixed.

\begin{algorithm}[!htb]
	\caption{Algorithm of CNTS.}
	\begin{algorithmic}[1]
		\REQUIRE
		The input data $\bold{W}$;
		Initialize the model \textbf{R} with parameters $\theta_R$;
		Initialize the model \textbf{D} with parameters $\theta_D$;
		Train epoch of CNTS epochs;
		Train epoch of \textbf{R} R\_epochs;
		Train epoch of \textbf{D} D\_epochs;
		\ENSURE
		The trained models \textbf{R} and \textbf{D} with parameters $\theta_R^*$ and $\theta_D^*$, respectively.;
		\STATE \textbf{for} e \textbf{in} epochs:
		\label{epoch}
		\STATE ~~~~\# Train phase of \textbf{R}.
		\STATE ~~~~\textbf{for} r\_e \textbf{in} R\_epochs:
		\STATE ~~~~~~~~$\bold{W_r} = R(\bold{W}, \theta_R)$;
		\STATE ~~~~~~~~$\bold{W_{r_s}}$ and $\bold{W_s}$ are selected by $\bold{W_r}$ and $\bold{W}$, \\respectively;
		\STATE ~~~~~~~~$\mathcal{L}_R = MSE(\bold{W_{r_s}}, \bold{W_s})$;
		\STATE ~~~~~~~~Update parameters of \textbf{R} to minimize the loss $\mathcal{L}_R$;
		\STATE ~~~~\textbf{end for}
		\STATE ~~~~\# Train phase of \textbf{D}.
		\STATE ~~~~\textbf{for} d\_e \textbf{in} D\_epochs:
		\STATE ~~~~~~~~$\bold{e_r} = MSE(\bold{W}, \bold{R}(\bold{W}, \theta_R))$;
		\STATE ~~~~~~~~$Score = \bold{D}(\bold{W}, \theta_D)$;
		\STATE ~~~~~~~~$Score_s$ and $\bold{e_{r_s}}$ are selected by $Score$ and $\bold{e_{r}}$,\\ respectively;
		\STATE ~~~~~~~~$\mathcal{L}_D = CrossEntropy(Softmax(Score_s),$\\ ~~~~~~~~~~~~~~~~~~~~~~~~~~~~~~~~~~~~~~~~~~~~~~~~$Softmax(\bold{e_{r_s}}))$;
		\STATE ~~~~~~~~Update parameters of \textbf{D} to minimize the loss $\mathcal{L}_D$;
		\STATE ~~~~\textbf{end for}
		\STATE \textbf{end for}
	\end{algorithmic}
	\label{alg1}
\end{algorithm}

\section{Experiments}
\subsection{Models}
As the outstanding performance of FEDformer \cite{zhou2022fedformer} in time series forecasting tasks demonstrates its suitability for time series data, this paper selects it as the network architecture for \textbf{D} and \textbf{R}. This study compares nine deep learning methods and presents them as follows:

\begin{itemize}
	\item {MERLIN \cite{nakamura2020merlin}}: an algorithm that can efficiently and
	exactly find discords of all lengths in massive time series archives.
	\item {LSTM-NDT \cite{hundman2018detecting}}: a long short-term memory network that effectively solves the problem of unsupervised anomaly detection.
	\item {OmniAnomaly \cite{su2019robust}}: a stochastic recurrent neural network for multivariate time series anomaly detection that works well robustly for various devices.
	\item {MSCRED \cite{zhang2019deep}}: a model that formulates the anomaly detection and diagnosis problem as three underlying tasks, i.e., anomaly detection,  root cause identification, and anomaly severity (duration) interpretation and addresses these issues jointly
	\item {MAD-GAN \cite{li2019mad}}: a framework considers the entire variable set concurrently to capture  the latent interactions amongst the variables.
	\item {USAD \cite{audibert2020usad}}: a multivariate time series anomaly detection model based on an autoencoder architecture whose learning is inspired by GANs.
	\item {MTAD-GAT \cite{zhao2020multivariate}}: a self-supervised framework for multivariate time-series anomaly detection 
	\item {CAE-M \cite{zhang2021unsupervised}}: a unsupervised deep learning based anomaly detection approach for multi-sensor time series data with two main subnetworks: characterization network and memory network.
	\item {GDN \cite{deng2021graph}}: a structure learning approach with graph neural networks, additionally using attention weights to provide explainability for the detected anomalies.
\end{itemize}

\subsection{Datasets}
The final model comparison in this study utilizes the following two datasets:

\begin{itemize}
	\item {NAB \cite{ahmad2017unsupervised}}: This dataset consists of labeled real-world and artificial time series, including metrics from AWS servers metrics, online ad click-through rates, real-time traffic data, and a collection of Twitter mentions from large public companies.
	\item {NASA-SMAP and NASA-MSL \cite{benecki2021detecting}}: These datasets consist of real spacecraft telemetry data from the Soil Moisture Active Passive (SMAP) satellite and the Curiosity rover (MSL), respectively. 
\end{itemize}

For the purposes of this study, only the first dimension representing continuous data is retained, while the remaining dimensions for binary data are omitted. In the experiments, the data is divided into a training set and a test set. The training set is used for model training and does not include abnormal labels, while the test set contains abnormal labels for comparison of results. Data without anomalies in the test set is removed. Finally, the effective data information is obtained, as shown in Table \ref{dataset}. The header of Table \ref{dataset} indicates the name of the datasets, the number of points in the training set, the number of points in the test set, the number of abnormal points in the test set, and the abnormal proportion.

\begin{table}[!htb]
	\caption{Information of Datasets}
	\centering
	\begin{tabular}{cccccc} 
		\hline
		Name      & Nums & \begin{tabular}[c]{@{}c@{}}Train\\Nums\end{tabular} & \begin{tabular}[c]{@{}c@{}}Test\\Nums\end{tabular} & \begin{tabular}[c]{@{}c@{}}Abnormal\\Nums\end{tabular} & \begin{tabular}[c]{@{}c@{}}Abnormal\\Rate (\%)\end{tabular}  \\ 
		\hline
		NAB       & 47   & 164096                                              & 164130                                             & 18456                                                  & 11.24                                                        \\
		NASA-MSL  & 27   & 58317                                               & 73729                                              & 7766                                                   & 10.53                                                        \\
		NASA-SMAP & 54   & 138004                                              & 435826                                             & 55817                                                  & 12.81                                                        \\
		Total     & 128  & 360417                                              & 673685                                             & 82039                                                  & 11.53                                                  \\
		\hline
	\end{tabular}
	\label{dataset}
\end{table}

\subsection{Detection}
In the experiments, anomaly detection is performed on all data separately, and the average value of each dataset is finally calculated. During the test phase, the data is fed into \textbf{D} to obtain the anomaly scores and the anomaly labels are used to calculate performance metrics. However, due to the overestimation of the anomaly detection model by the anomaly adjustment strategy \cite{xu2018unsupervised}, the detection results reported in many previous studies are no longer credible \cite{kim2022towards}. Hence, this paper eliminates the anomaly adjustment strategy and utilizes a unified detection metric to compare different methods. This metric calculates the result by combining all the test data, instead of adjusting for anomalies. A point with an anomaly score greater than a selected threshold $\delta$ is considered an anomaly. The output anomaly result $\bold{S^{\delta}} = \{s_1^{\delta}, s_2^{\delta}, \dots, s_N^{\delta}\}$ is the following formula
\begin{equation}
	s_i^{\delta} = 
	\begin{cases}
		1 \quad if\ Score(\bold{x_i}) > \delta,\\
		0 \quad else.
	\end{cases}
\end{equation}
where $Score(\bold{x_i})$ is the anomaly score of $\bold{x_i}$. After labeling, the accuracy, precision, recall, and F1 score for the evaluation are computed as follows:
\begin{equation}
	acc = \frac{TP + TN}{TP + FP + FN + TN}
\end{equation}
\begin{equation}
	precision = \frac{TP}{TP + FP}
\end{equation}
\begin{equation}
	recall = \frac{TP}{TP + FN}
\end{equation}
\begin{equation}
	F1 = \frac{2 \cdot precision \cdot recall}{precision + recall}
\end{equation}
where TN, TP, FP, and FN denote the number of true negatives, true positives, false positives and false negatives, respectively.

There are numerous ways to choose the threshold \cite{siffer2017anomaly, tibshirani2001estimating}. In this paper, the focus is on evaluating the model's ability to distinguish between abnormal and normal data, therefore, the threshold that maximizes the F1 score of the model results is selected.
\begin{equation}
	\delta^* = arg \underset{\delta}{max}\ F1(\bold{Y}, \bold{S^{\delta}})
\end{equation}
where $F1(\bold{Y}, \bold{S^{\delta}})$ is F1 score between $\bold{Y}$ and $\bold{S^{\delta}}$. The accuracy (ACC), F1 value and area under the curve (AUC) of receiver operating characteristic (ROC) of the detection results are shown in the table \ref{detection}. 

\begin{table*}[!htb]
	
	\centering
	\caption{Comparison of Detection Results}
	\begin{threeparttable}  
		\begin{tabular}{l|lll|lll|lll} 
			\hline
			\multicolumn{1}{c}{} & \multicolumn{3}{c}{NAB}                                                    & \multicolumn{3}{c}{NASA-MSL}                                               & \multicolumn{3}{c}{NASA-SMAP}                                               \\ 
			\cline{2-10}
			\multicolumn{1}{c}{} & \multicolumn{1}{c}{ACC} & \multicolumn{1}{c}{F1} & \multicolumn{1}{c}{AUC} & \multicolumn{1}{c}{ACC} & \multicolumn{1}{c}{F1} & \multicolumn{1}{c}{AUC} & \multicolumn{1}{c}{ACC} & \multicolumn{1}{c}{F1} & \multicolumn{1}{c}{AUC}  \\ 
			\hline
			TranAD               & 0.5180                   & 0.2989                 & 0.5818                  & 0.7820                   & 0.4587                 & 0.6765                  & 0.7775                  & 0.4526                 & 0.7133                   \\
			GDN                  & 0.5665                  & 0.2909                 & 0.5785                  & 0.8111                  & 0.4578                 & 0.6871                  & 0.8050                   & 0.4207                 & 0.7022                   \\
			MAD\_GAN             & 0.5161                  & 0.2944                 & 0.5781                  & 0.8099                  & 0.4503                 & 0.6816                  & 0.8043                  & 0.4251                 & 0.7019                   \\
			MTAD\_GAT            & 0.5611                  & 0.2978                 & 0.5873                  & 0.7730                   & 0.4411                 & 0.6780                   & 0.8146                  & 0.4030                  & 0.6951                   \\
			MSCRED               & 0.4940                   & 0.2845                 & 0.5727                  & 0.7709                  & 0.4407                 & 0.6791                  & 0.7613                  & 0.3982                 & 0.6788                   \\
			USAD                 & 0.5383                  & 0.2952                 & 0.5823                  & 0.7954                  & 0.4489                 & 0.6825                  & 0.7954                  & 0.4161                 & 0.6989                   \\
			OmniAnomaly          & 0.5348                  & 0.2926                 & 0.5768                  & 0.8126                  & 0.4305                 & 0.6704                  & 0.7479                  & 0.4048                 & 0.6847                   \\
			LSTM\_AD             & 0.4887                  & 0.2873                 & 0.5746                  & 0.6916                  & 0.4065                 & 0.6603                  & 0.6297                  & 0.3751                 & 0.6460                    \\
			CAE\_M               & 0.5504                  & 0.2941                 & 0.5828                  & 0.7605                  & 0.4400                   & 0.6720                   & 0.8033                  & 0.4139                 & 0.7003                   \\
			Detection\tnote{*}       & 0.5937                  & 0.3177                 & 0.6071                  & 0.7555                  & 0.4715                 & 0.7109                  & 0.7499                  & 0.4852                 & 0.6682                   \\
			CNTS                 & \textbf{0.7010}          & \textbf{0.3717}        & \textbf{0.6350}          & \textbf{0.8277}         & \textbf{0.5459}        & \textbf{0.7320}          & \textbf{0.8733}         & \textbf{0.5685}        & \textbf{0.7396}          \\
			\hline
		\end{tabular}
		\begin{tablenotes}   
			\footnotesize              
			\item[*] This method is a network structure that takes out the detector separately in CNTS and uses the reconstruction loss as the training target.
		\end{tablenotes}            
	\end{threeparttable} 
	\label{detection}
\end{table*}

The results of the comparison between different detection methods are depicted in Table \ref{detection}, with the optimal result highlighted in bold. It is evident that the CNTS outperforms the other methods on all three datasets after removing the influence of the abnormal adjustment. This confirms the previous finding that the abnormal adjustment tends to overestimate the performance of the model \cite{kim2022towards, hwang2022you}. The table shows that the CNTS has greatly improved, resulting in the best results in terms of ACC, F1, and AUC. Compared to the baseline method, there has been an improvement of 15\%, 17\%, and 6\% respectively in these indicators. Additionally, it has been demonstrated that the use of FEDformer alone, as an unsupervised anomaly detection model based on reconstruction, can still produce good results for some indicators. This indicates a degree of correlation across different time series data tasks.

\subsection{Reconstruction}
The obtained reconstructed results form \textbf{R} and the original data are used to calculate the metrics. The reconstruction error $\bold{e_r} = \{e_1, e_2, \dots, e_N\}$.The reconstruction error of normal and abnormal values will be used to measure the quality of the reconstructed model. It is calculated as follows:
\begin{equation}
	\begin{aligned}
		M&SE_n = \frac{\sum_{i=0}^{N} e_i * (1 - y_i)}{\sum_{i=0}^{N}({1 - y_i})} \\
		&MSE_a = \frac{\sum_{i=0}^{N} e_i * y_i}{\sum_{i=0}^{N}y_i}
	\end{aligned}
\end{equation}
where $MSE_n$ and $MSE_a$ represent the MSE errors of normal and abnormal, respectively.

\begin{figure}[!htb]
	\centering
	\includegraphics[width=\linewidth]{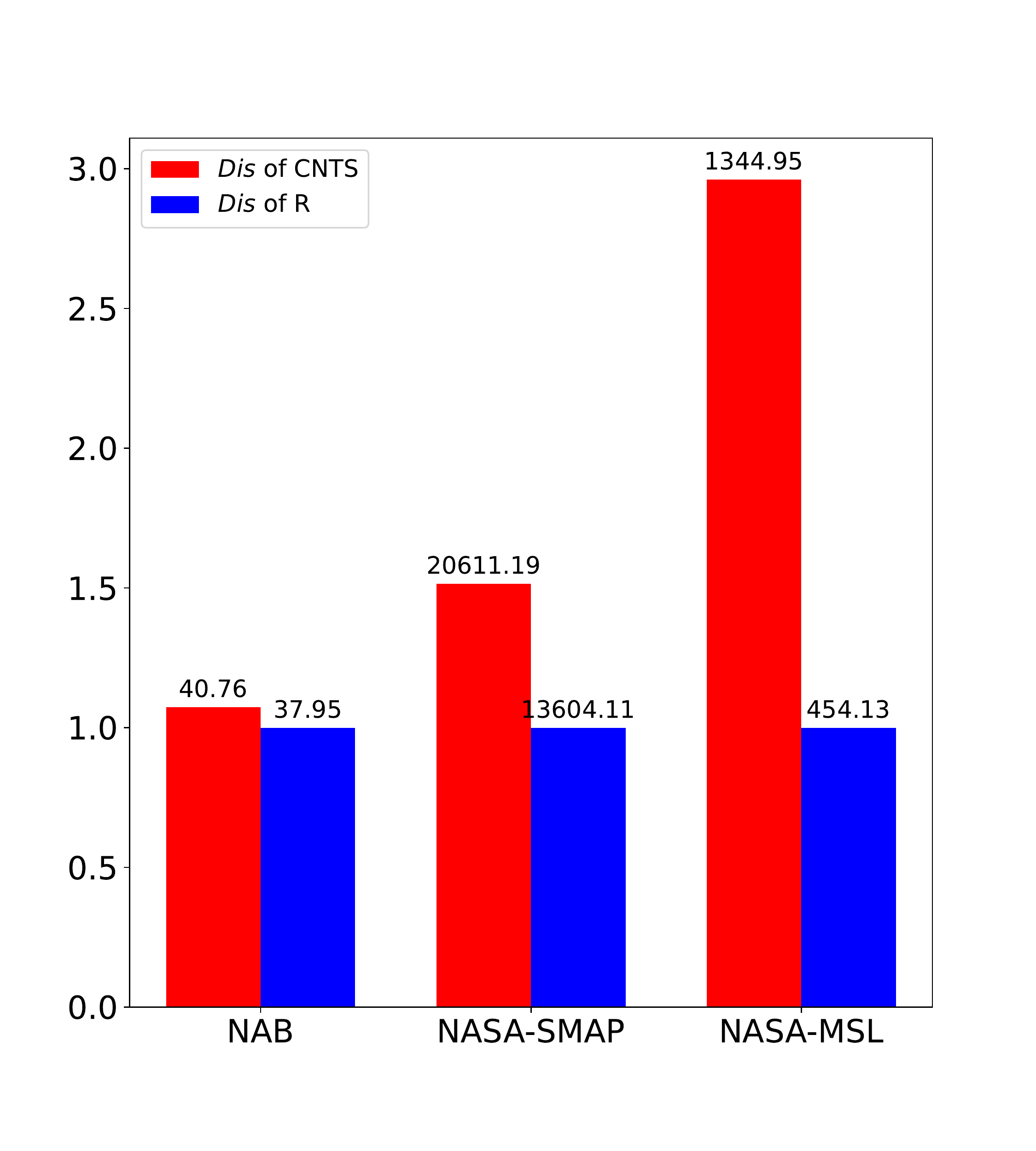}
	\caption{Comparisons of reconstruction effects}
	\label{reconstruction}
\end{figure}

Figure \ref{reconstruction} presents a comparison of the reconstruction effectiveness between a single reconstruction network (\textbf{R}) and CNTS. The length of each bar in the histogram represents the proportionality of the models, not the actual values, with \textbf{R} being equal to 1. The numerical value for each model's index is indicated on the top of each bar. The histogram further quantifies the capability of the models in distinguishing between normal and anomalous data points. This value can be calculated using the formula provided.

\begin{equation}
	Dis = \frac{MSE_{a} - MSE_{n}}{MSE_{n}}
\end{equation}

The histogram in Figure \ref{reconstruction} clearly illustrates the superiority of CNTS over the baseline model in terms of discriminative power. As can be observed from the figure, the $Dis$ value of CNTS is higher than that of the baseline model across all datasets, with a margin of approximately three times higher on the NASA-MSL data. This result further emphasizes the detrimental impact of outliers on the training process of reconstruction models, reducing their sensitivity to anomalies.

\subsection{Cooperation}
\begin{figure*}[!htb]
	\centering
	
	\subfloat[A-4 in NASA-SMAP]{
		\includegraphics[width=2.3in]{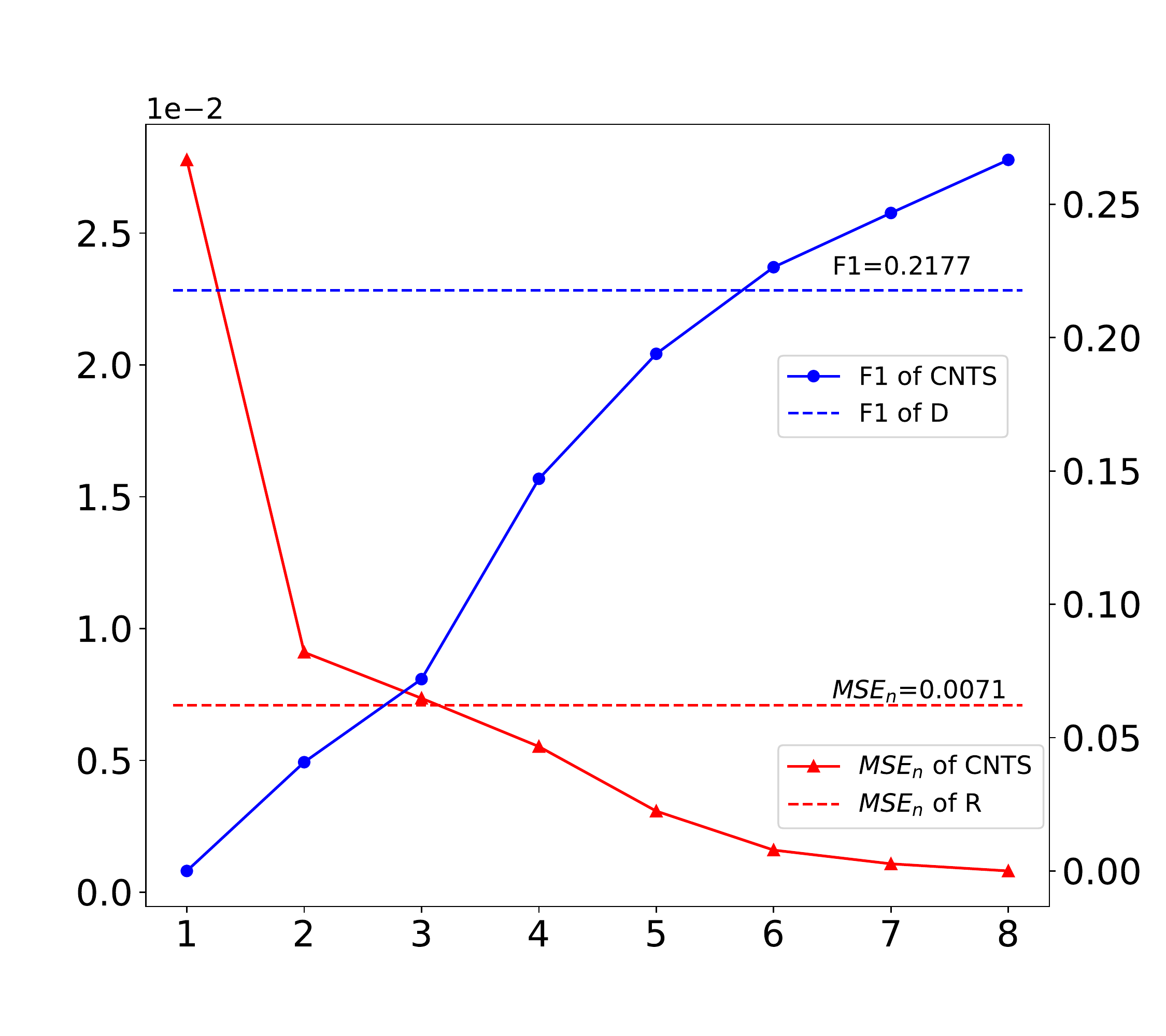}
	}
	\subfloat[D-14 in NASA-MSL]{
		\includegraphics[width=2.3in]{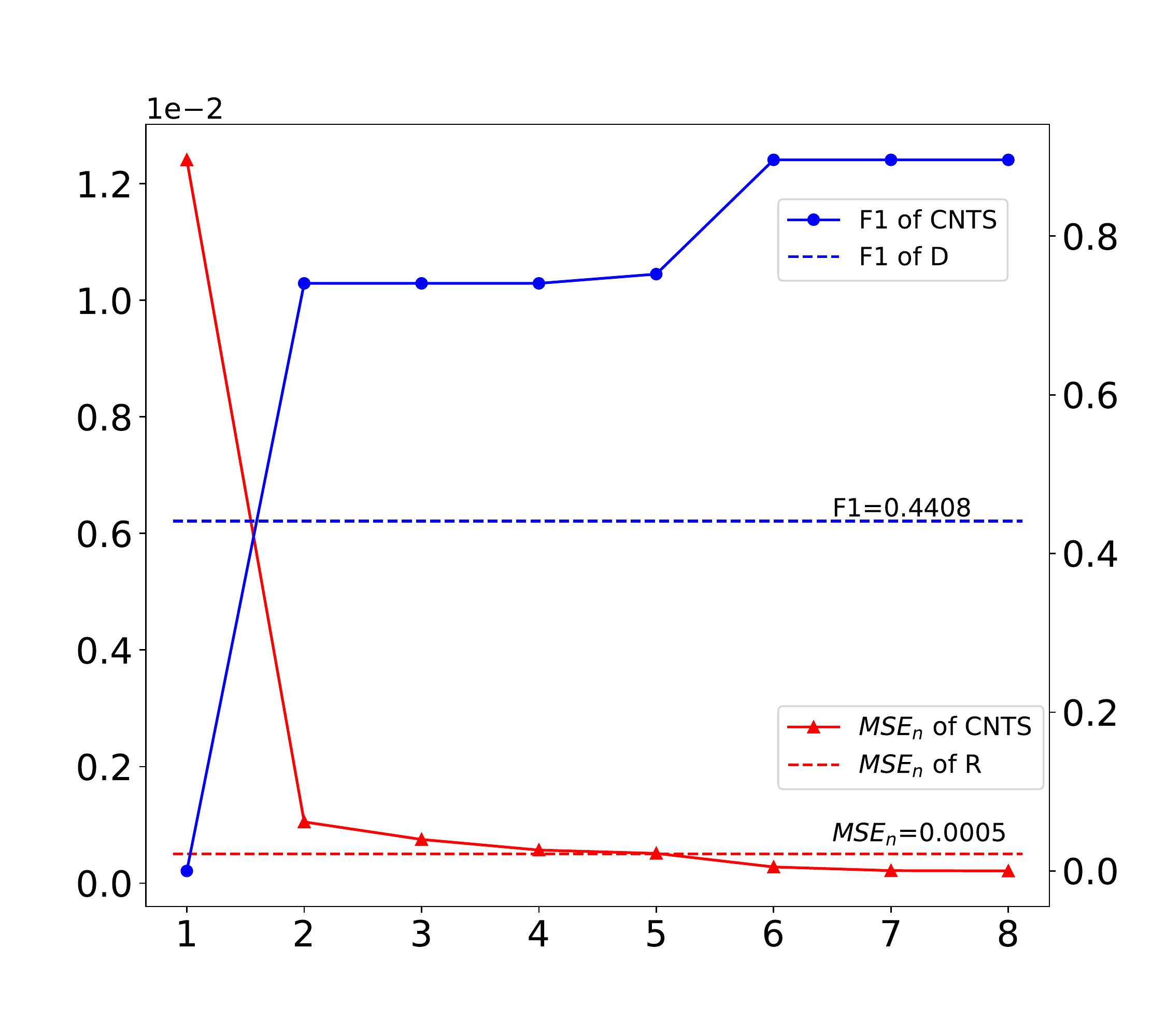}
	}%
	\subfloat[CloudWatch 13 in NAB]{
		\includegraphics[width=2.3in]{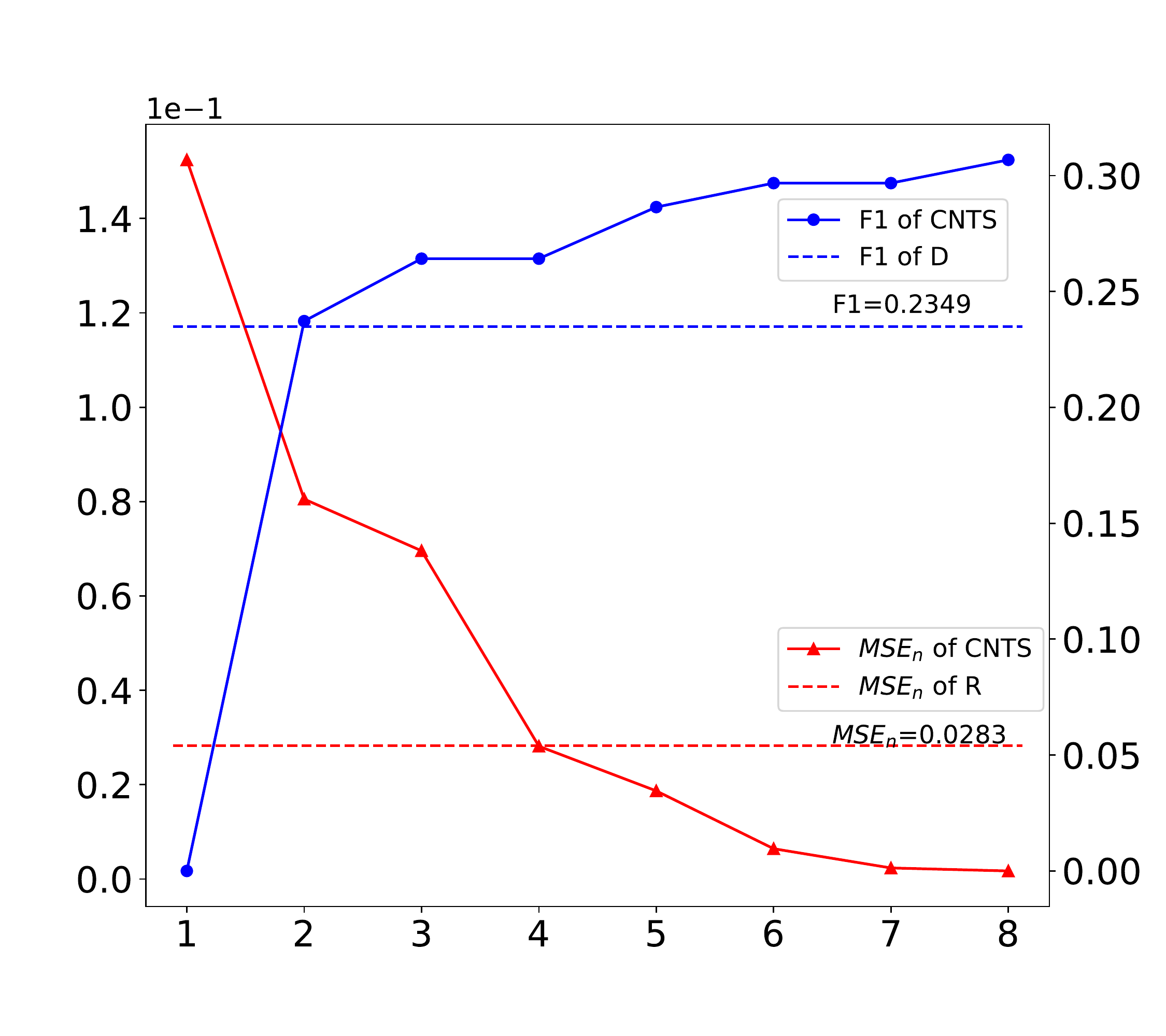}
	}%
	\hfil
	\subfloat[G-7 in NASA-SMAP]{
		\includegraphics[width=2.3in]{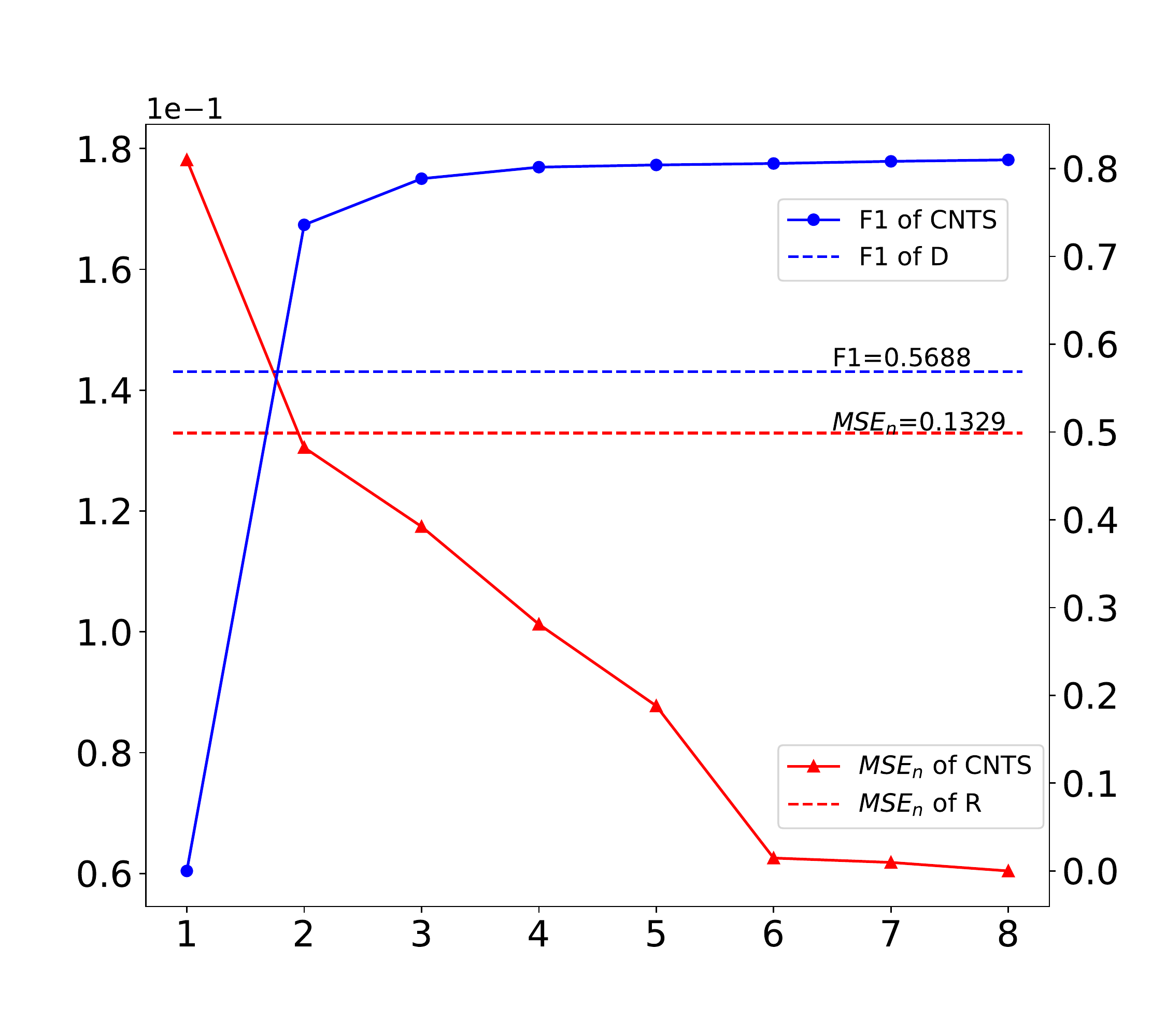}
	}%
	\subfloat[F7 in NASA-MSL]{
		\includegraphics[width=2.3in]{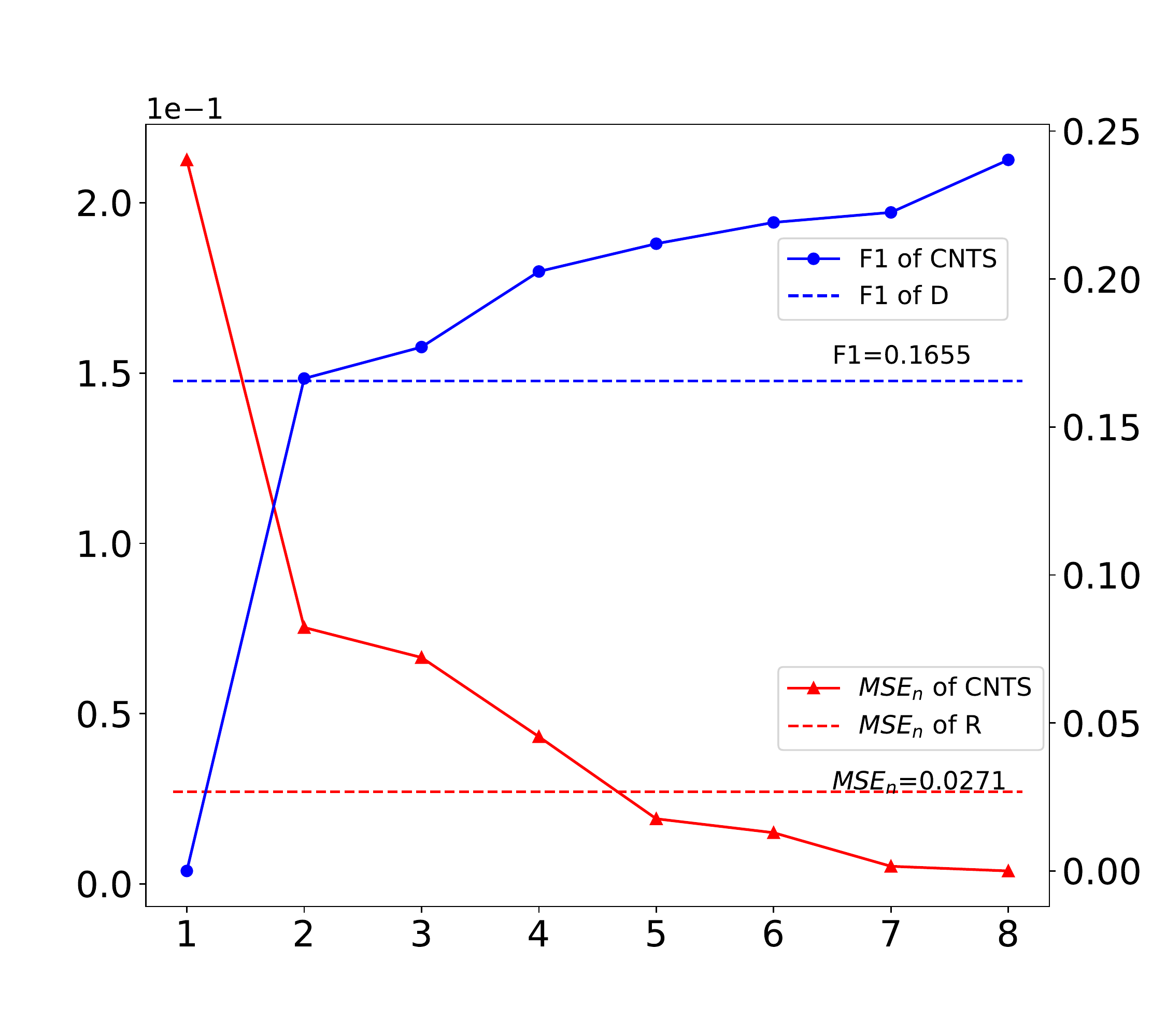}
	}%
	\subfloat[Exchange 5 in NAB]{
		\includegraphics[width=2.3in]{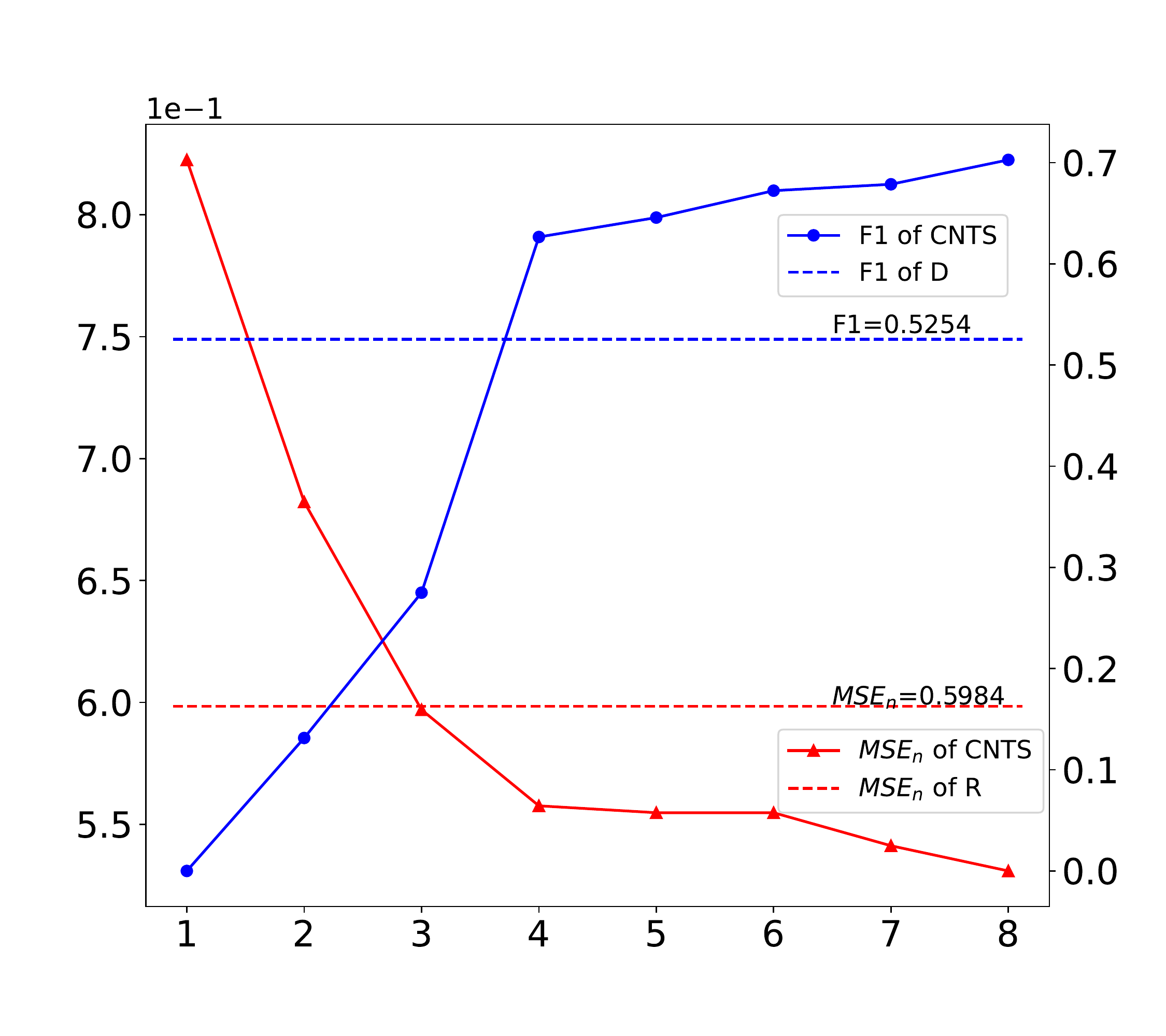}
	}%
	\hfil
	\caption{Metric Changes of Cooperative Training. The solid lines represent the results of CNTS, and the dashed lines represent the results of single \textbf{D} or \textbf{R}.}
	\label{stages}
\end{figure*}

Figure \ref{stages} displays the training process for six data points from three different databases. The x-axis in Figure \ref{stages} represents the number of alternating training rounds, while the left and right y-axes represent the reconstruction loss for normal data points and the F1 value of the detector, respectively.

As the training progresses, the reconstruction loss of CNTS for normal values continually decreases, thereby exposing outliers to an increasingly higher reconstruction error. As a result, the detection component of CNTS continuously learns from this reconstruction error to enhance its ability to detect anomalies.

By combining these results, it can be observed that CNTS outperforms the baseline method after one stage of training in some data sets, while in others it may require several stages of training to gradually surpass the baseline method. Regardless of the starting point, as the training phase advances, the performance of CNTS's detector and reconstructor continues to improve. This demonstrates that there is a cooperative relationship between the detector and reconstructor during the training process, resulting in superior performance compared to training them individually.

\begin{figure*}[!htb]
	\centering
	
	\subfloat[A-4 in NASA-SMAP]{
		\includegraphics[width=2.3in]{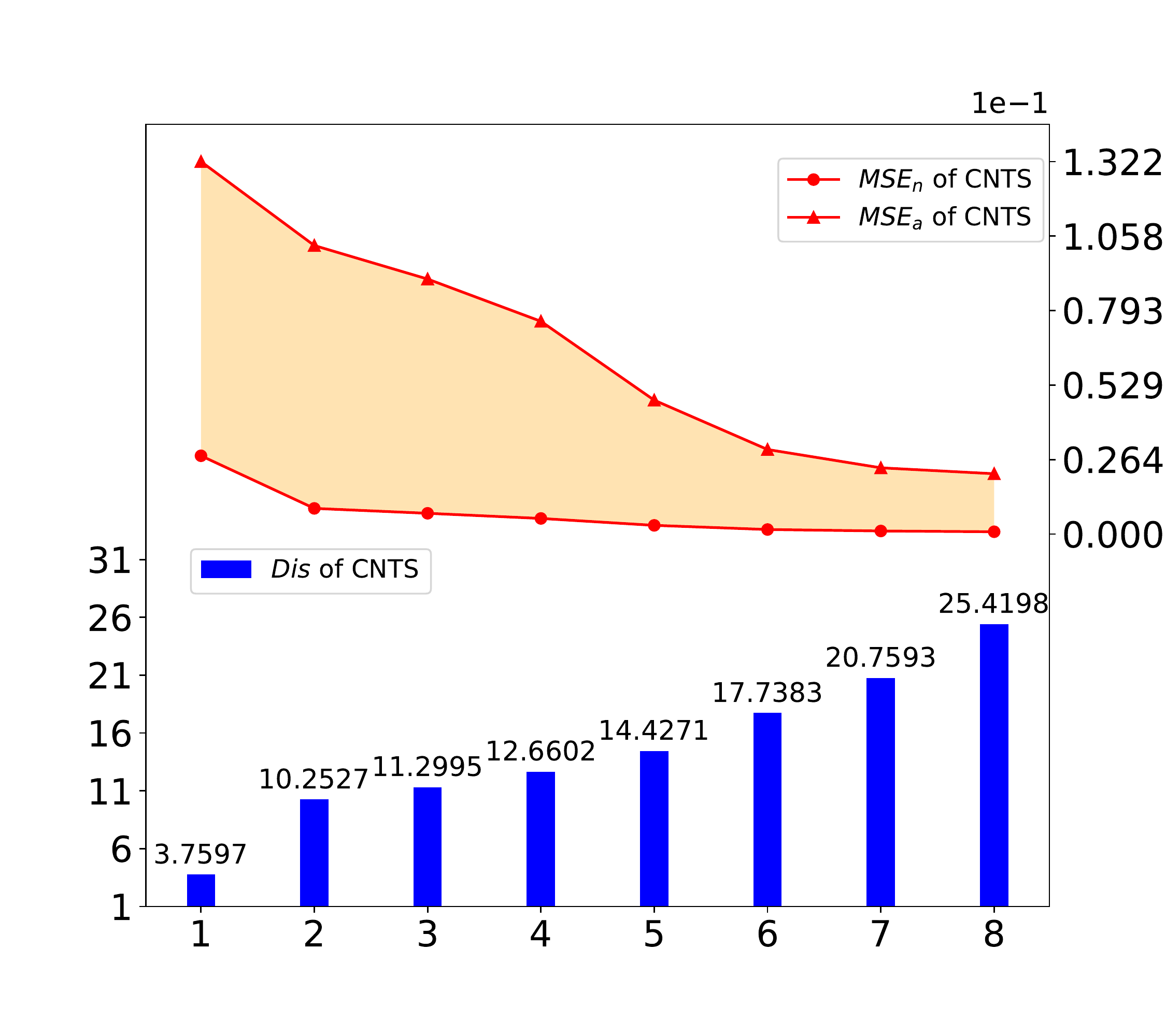}
	}%
	\subfloat[D-14 in NASA-MSL]{
		\includegraphics[width=2.3in]{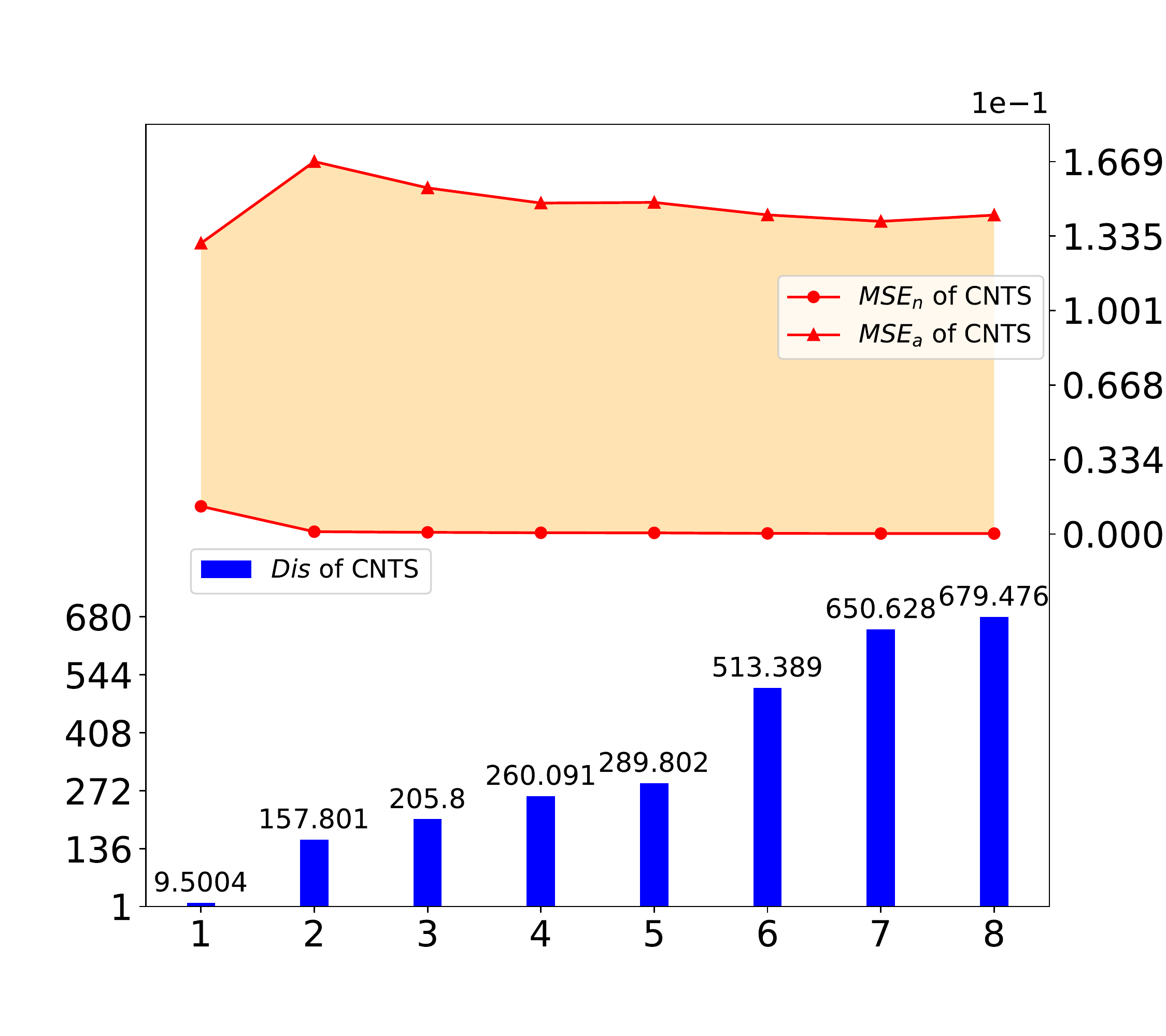}
	}%
	\subfloat[CloudWatch 13 in NAB]{
		\includegraphics[width=2.3in]{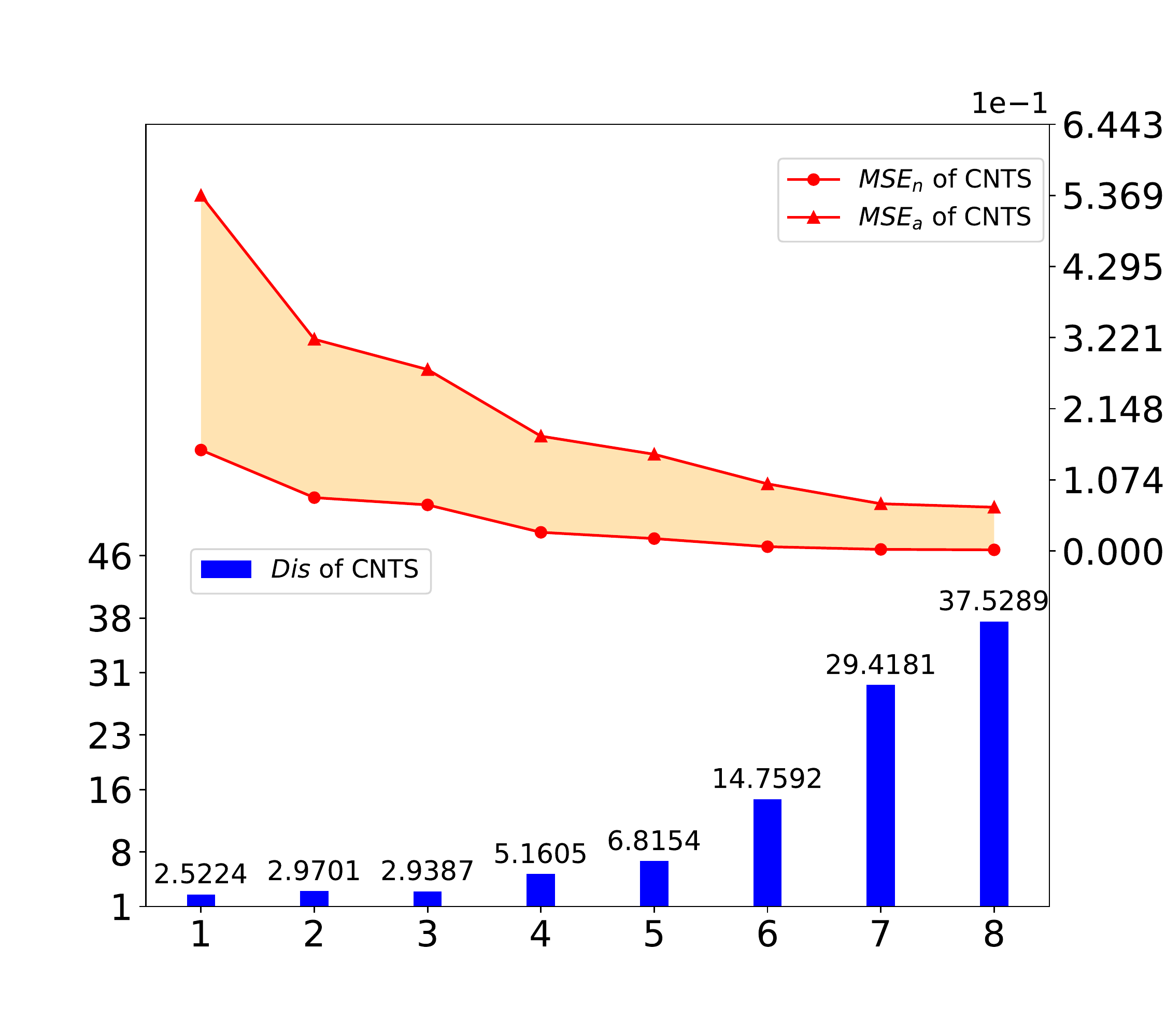}
	}%
	\hfil
	\subfloat[G-7 in NASA-SMAP]{
		\includegraphics[width=2.3in]{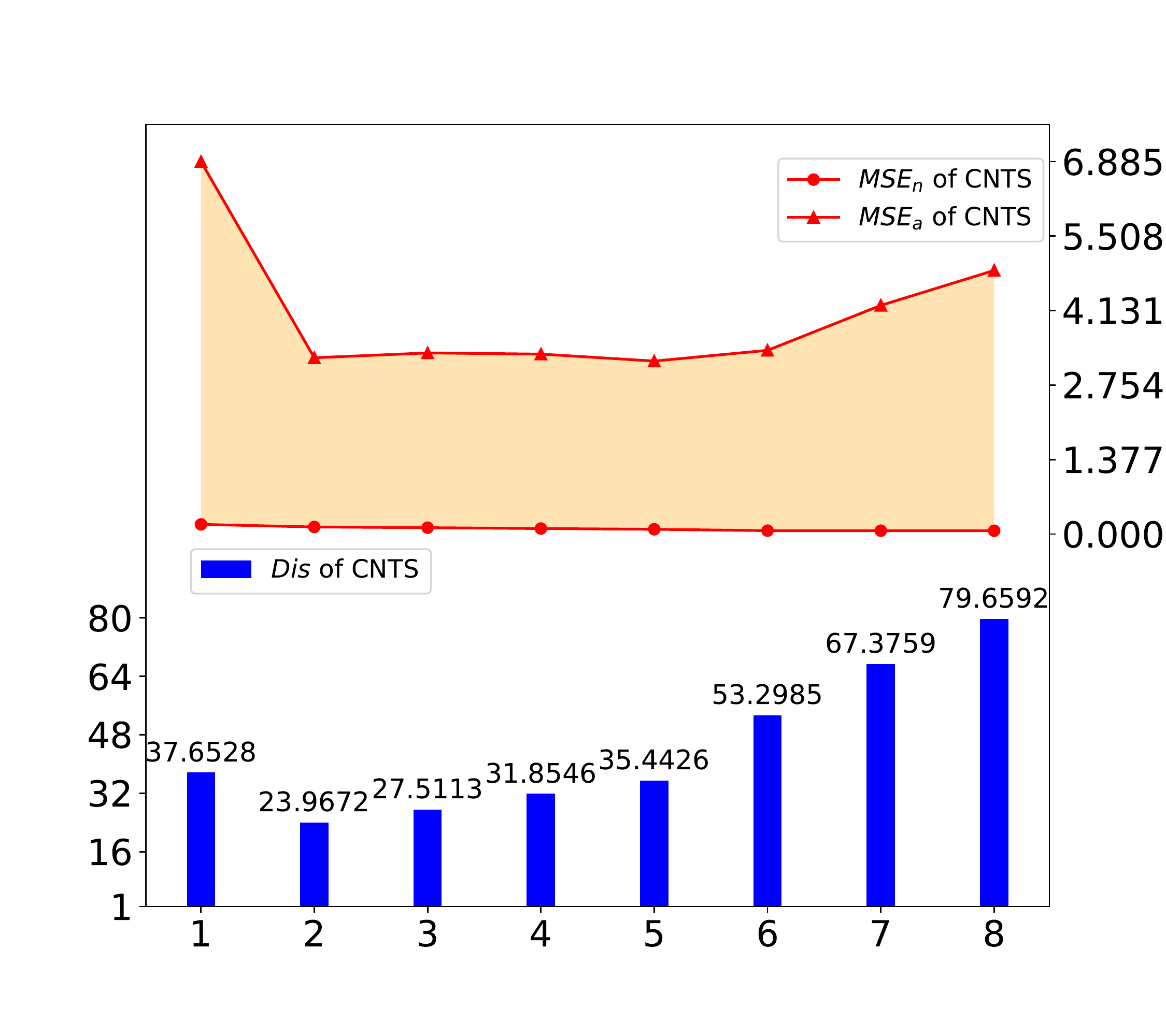}
	}%
	\subfloat[F7 in NASA-MSL]{
		\includegraphics[width=2.3in]{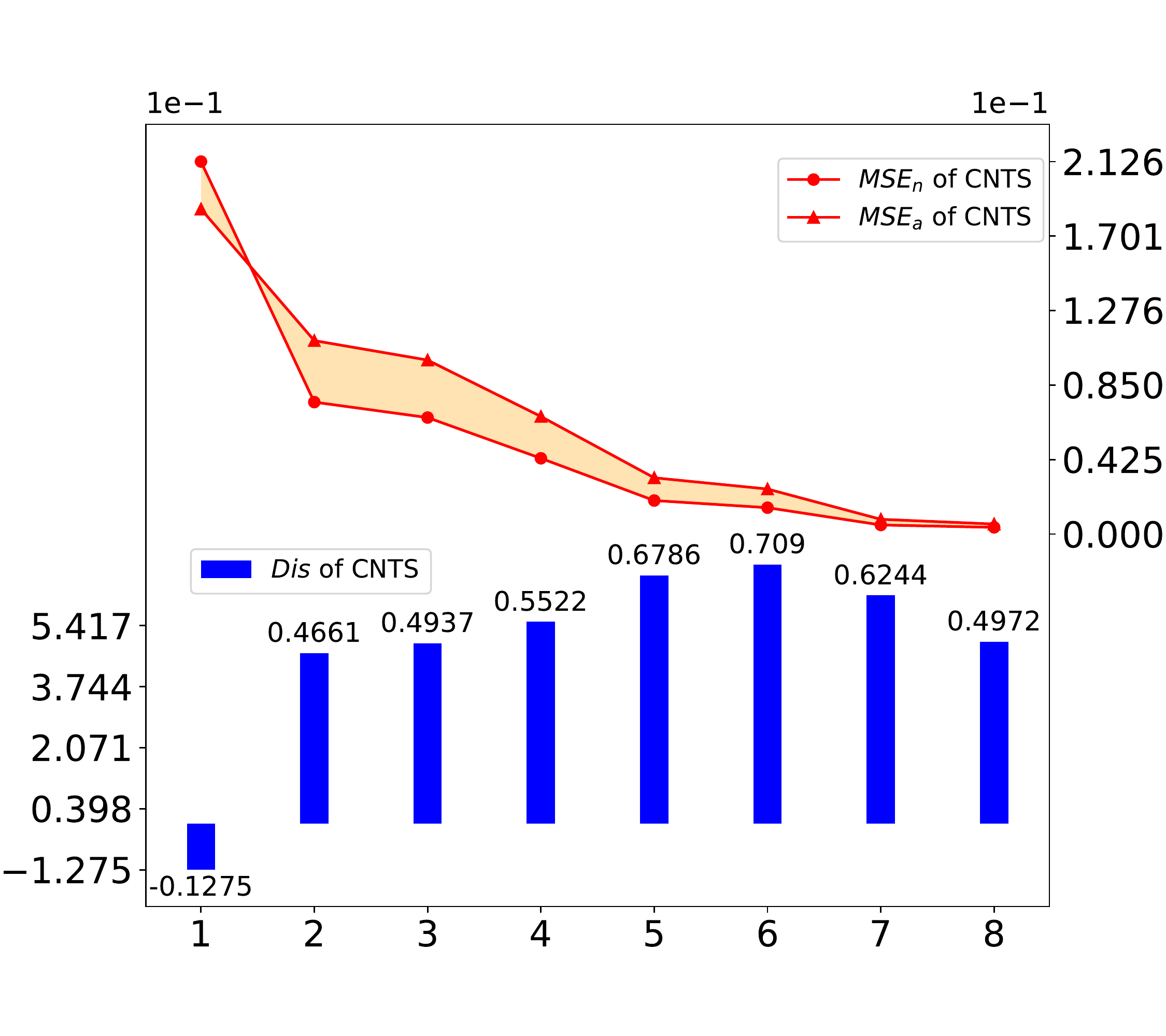}
	}%
	\subfloat[Exchange 5 in NAB]{
		\includegraphics[width=2.3in]{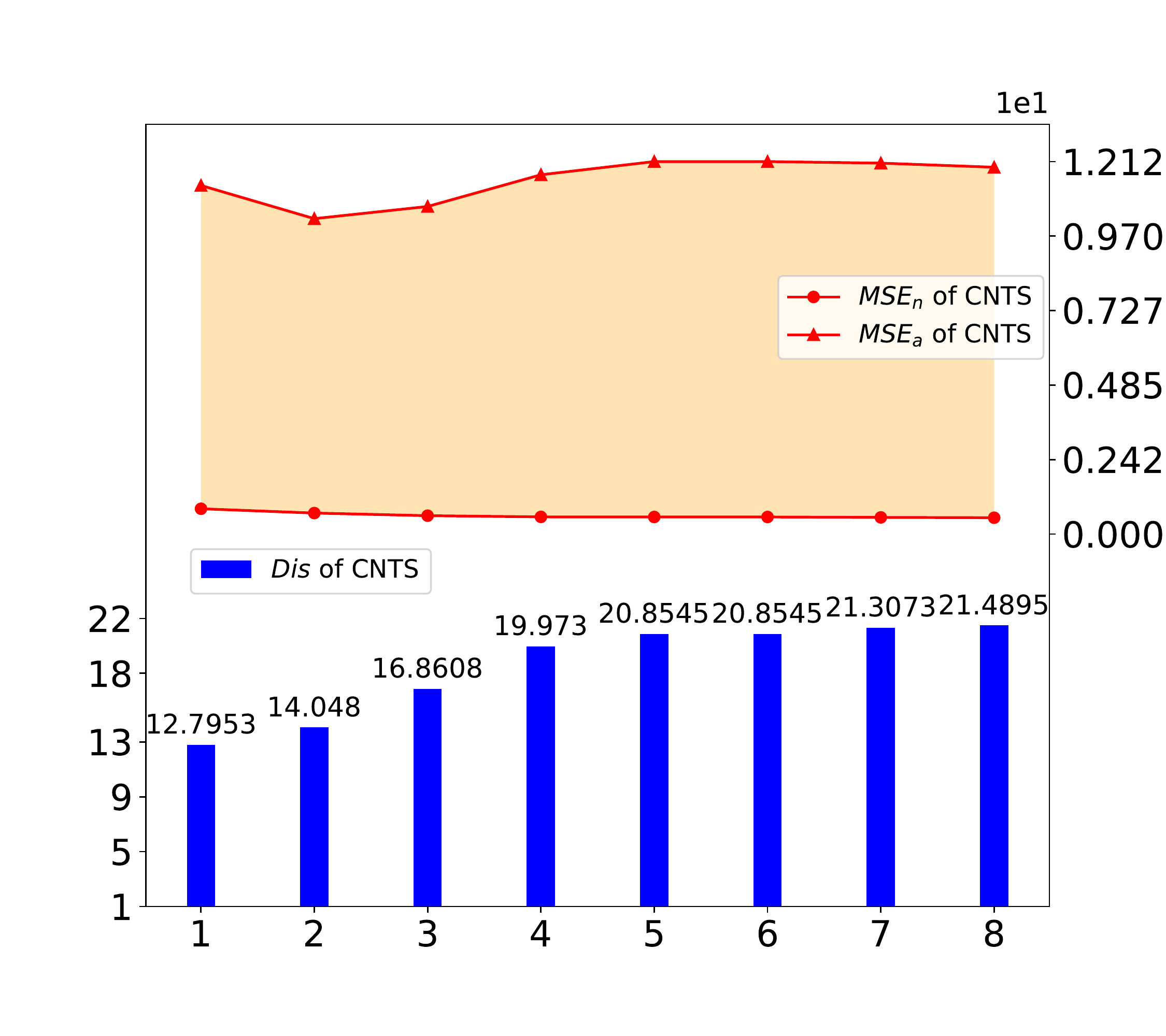}
	}%
	\hfil
	\caption{Metric Changes of Cooperative Training. The red line represents the reconstruction error, with triangles and dots representing the reconstruction errors of outliers and normal values, respectively. The histograms depict the $Dis$ of the model at different stages of training.}
	\label{stages_rec}
\end{figure*}

Figure \ref{stages_rec} illustrates the variations in various indicators during the training process of CNTS's \textbf{R} as the stage advances. The red line represents the reconstruction error, with triangles and dots representing the reconstruction errors of outliers and normal values, respectively. The histograms depict the $Dis$ of the model at different stages of training. 

The line graph reveals that the reconstruction errors of normal values consistently decrease, demonstrating that the model is learning effectively. However, the reconstruction errors of outliers are more fluctuating. The histograms provide another perspective, showing that \textbf{R} has the ability to differentiate between outliers and normal values. Although there may have been some fluctuations in earlier stages, the overall trend indicates that the $Dis$ is increasing, which implies that the discriminative ability of \textbf{R} improves as the number of model training increases.

\section{Conclusion}
In this paper, we propose a novel cooperation-based unsupervised anomaly detection method for time series data, referred to as CNTS. This method is based on reconstruction loss and aims to address the limitations of existing reconstruction-based unsupervised anomaly detection methods by mitigating the impact of outliers on reconstruction performance. CNTS integrates a separate network dedicated to learning the anomaly detection task. The experimental results demonstrate that CNTS effectively helps the reconstruction model to distinguish between normal and abnormal time series data. 

There are three aspects that can be further studied in the CNTS. One area of focus could be the selection of the basic model, such as FEDformer. Currently, it is not specifically designed for anomaly detection or reconstruction, and it may be beneficial to investigate the use of a dedicated network for these tasks. Additionally, to reduce the time complexity of CNTS, which results from its dual-network structure, one can alleviate it by incorporating a more efficient approach, such as missing value imputation, prediction, or classification, into a unified framework for processing time series data. Finally, further experimentation is useful to determine the optimal proportion of data to be selected or removed during the data selection process by detectors and reconstructors.


\bibliographystyle{IEEEtran}
\bibliography{CNTS}

\begin{IEEEbiography}[{\includegraphics[width=1in,height=1.25in,clip,keepaspectratio]{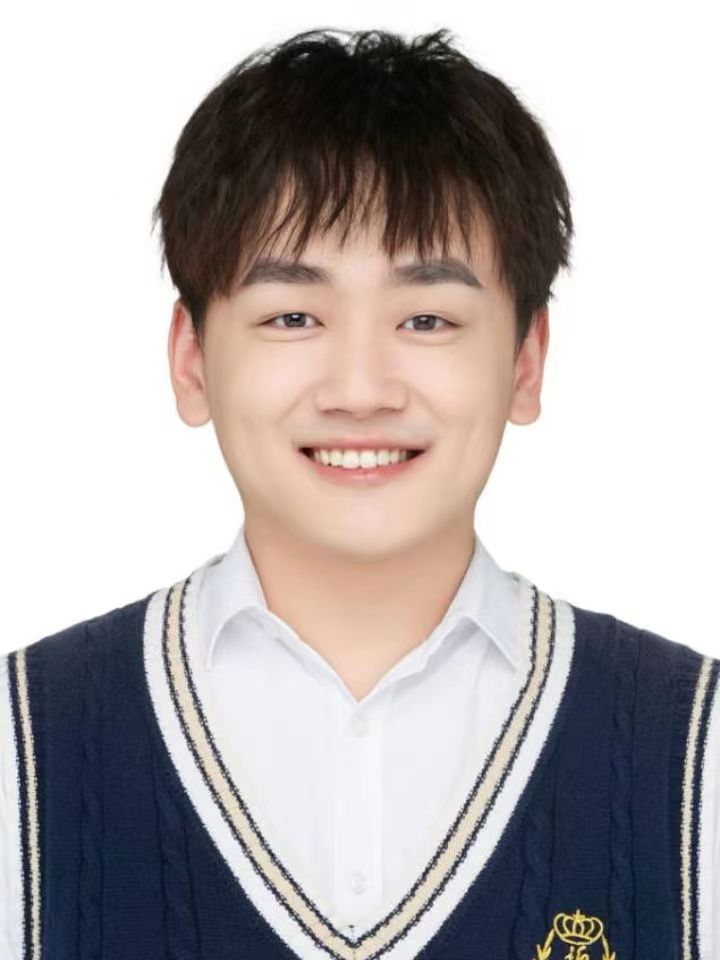}}]{Jinsheng Yang}
	received a bachelor's degree from Sichuan University in 2016. Currently, he is a postgraduate student at Hainan University. His main research direction is time series data analysis.
\end{IEEEbiography}

\begin{IEEEbiography}[{\includegraphics[width=1in,height=1.25in,clip,keepaspectratio]{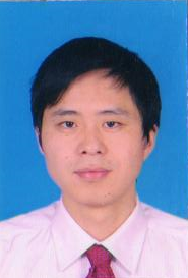}}]{Yuanhai Shao} 
	received his B.S. degree in information and computing science from College of Mathematics, Jilin University, a master's degree in applied mathematics, and a Ph.D. degree in operations research and management in College of Science from China Agricultural University, China, in 2006, 2008 and 2011, respectively. Currently, he is a Full Professor at the School of Management, Hainan University, Haikou, China. His research interests include support vector machines, optimization methods, machine learning and data mining. He has published over 100 refereed papers on these areas, including IEEE TPAMI, IEEE TNNLS, IEEE TC, PR, and NN.
\end{IEEEbiography}

\begin{IEEEbiography}[{\includegraphics[width=1in,height=1.25in,clip,keepaspectratio]{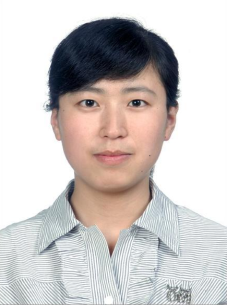}}]{Chun-Na Li}
	received her Master's degree and Ph.D degree in Department of Mathematics from Harbin Institute of Technology, China, in 2009 and 2012, respectively. Currently, she is a professor at Management School, Hainan University. Her research interests include optimization methods, machine learning and data mining.
\end{IEEEbiography}

\EOD

\end{document}